\documentclass{article}
\usepackage{spconf,amsmath,epsfig}
\usepackage{graphicx}
\usepackage{url}

\title{Artificial Generation of Big Data For Improving Image Classification:\\ A Generative Adversarial Network Approach on SAR Data}

\name{
  \parbox{\linewidth}{\centering	
  *Dimitrios Marmanis\textsuperscript{1,3},
  *Wei Yao\textsuperscript{1},
  Fathalrahman Adam\textsuperscript{1},
  Mihai Datcu\textsuperscript{1}, \\
  \textit{Peter Reinartz}\textsuperscript{1},
  \textit{Konrad Schindler}\textsuperscript{2},
  \textit{Jan Dirk Wegner}\textsuperscript{2},
  \textit{Uwe Stilla}\textsuperscript{3}
  \thanks{*Authors have contributed equally in this work}
}}

\address{
  \textsuperscript{1}Department of Photogrammetry \& Image
  Analysis, German Aerospace Center (DLR), Germany \\
  \textsuperscript{2}Photogrammetry \& Remote Sensing Group, ETH Zurich,
  Switzerland \\
  \textsuperscript{3}Department Photogrammetry \&
  Remote Sensing, Technische Universitaet Muenchen (TUM), Germany
}

\begin{document}

\maketitle

\begin{abstract}

Very High Spatial Resolution (VHSR) large-scale \emph{SAR} image
databases are still an unresolved issue in the Remote Sensing field.
In this work, we propose such a dataset and use it to explore
patch-based classification in urban and peri-urban areas, considering
$7$ distinct semantic classes.  In this context, we investigate the
accuracy of large CNN classification models and pre-trained networks
for SAR imaging systems.  Furthermore, we propose a Generative
Adversarial Network (GAN) for SAR image generation and test, whether
the synthetic data can actually improve classification accuracy.

\end{abstract}
%


\section{Introduction}
\label{sec:intro}

Classification of very high resolution (VHR) \emph{SAR} image data
remains a hard and time-consuming task.
Major difficulties include the scarcity of available data, and the
challenge of semantically interpreting the \emph{SAR} backscatter
signal.
Linked to those difficulties, there are no large-scale,
\emph{SAR-derived} image databases for Remote Sensing image analysis
and knowledge discovery.
Furthermore, while optical image classification has seen a
breakthrough with the advent of \emph{Deep Learning} methods that
require Big Data, SAR-based systems have so far not experienced the
same progress, likely because of not enough data with associated
training labels is available.

In this work we try to tackle the lack of training data, by
introducing a large-scale \emph{SAR} image database.
Precisely, our dataset contains more than $60'000$ image instances and
respective labels, chosen from $7$ distinct semantic classes.
Using this data, we perform a set of experiments to understand the
impact of dataset size on classification accuracy.
In this context, we also investigate the possibility to further expand
the dataset with synthetic SAR images generated with the help
of\emph{Generative Adversarial Networks} (\emph{GANs}).
These are powerful generative models that have been shown to produce
high-quality synthetic images in other fields, thereby reducing (or
even compeltely avoiding) the annotation effort.
Our main contributions in this work can be summarized as follow:

\begin{itemize}

  \vspace{-0.33em}
  \item[$\bullet$] We construct the first state-of-the-art CNN model
    pre-trained on large-scale SAR data.

  \vspace{-0.33em}
  \item[$\bullet$] We investigate the possibility of transfer-learning
    from other pre-trained models based on optical images, and their
    impact on SAR image classification.

  \vspace{-0.33em}
  \item[$\bullet$] We investigate the possibility of training also
    with artificial SAR data generated with a GAN.

\end{itemize}

\section{Related Work}
\label{sec:previous work}

In the field of SAR image analysis, the use of deep-learning methods,
such as \emph{CNNs}, is still in its infancy, mainly due to the
limited availability of VHR data with asociated ground truth labels.
We note that, in a detailed literature review, we did not find any
work that relies on a large scale SAR-database to unlock the potential
of deep neural networks.
Moreover, there are no pre-trained networks for SAR images, which
would facilitate the classification of SAR datasets for which there
aren't enough training labels to learn a deep network from scratch.

Published work at the intersection of \emph{SAR} imaging and deep
learning are mainly focussed on \emph{Target Classification}. Some
representative works employ sparsely connected layers
\cite{chen2016target}, limited training data \cite{lin2017deep} and
domain-specific data augmentation methods
\cite{ding2016convolutional}.
%
%
In the field of \emph{GAN}s for \emph{SAR} data, some interesting
results have been shown by \cite{guo2017synthetic}, where authors
constructed a generative deep model.
The outcome of their experiments however remain unconclusive, due to
the scarcity of training data, and particular characteristics of the
underlying targets (military imagery).
Another implementation of \emph{GANs} in the field of Remote Sensing
is the one of \cite{perezsemi}, who investigate the \emph{Wasserstein
  GAN} for poverty mapping with sparse labels, using a semi-supervised
approach. They however do not use SAR imagery.
Yet another work on optical remote sensing imagery and artificial data
generation is the one of \cite{lin2016deep}. Thery propose an
additional objective function over the standard GAN architecture to
improve the output. While the approach is interesting, it ultimately
does not produce visually realistic images of the target classes.
A promising work is \cite{merkle2017possibility}, which demonstrates
the generation of synthetic SAR images on the basis of optical images.
The high-quality samples generated in that work show the potential of
\emph{GAN} methods for SAR image synthesis, and motivate us to further
investigate that topic.

\section{The Dataset}
\label{sec:dataset}

Our dataset was obtained via a novel classification scheme especially
designed for high-resolution SAR imagery of (mainly)q built-up areas.
The dataset contain image patches from $288$ TerraSAR-X image scenes
($41$ scenes acquired in Africa, $6$ from Antarctica, $59$ from Asia,
$80$ from Europe, $40$ from the Middle East, $54$ from North and South
America and $8$ from ocean surfaces), with a total of over $60'000$
individual patches.
All \emph{TerraSAR-X} data are obtained via the X-band instrument,
using the high-resolution Spotlight mode. The incident angles
throughout the scenes varies between 20 and 50 degrees. The resolution
of the images scenes is set to $2.9$m, with a pixel spacing of
$1.25$m.  The chosen polarization for the dataset is horizontal (HH)
for all products.  Furthermore, for conveniennce we convert all
intensity data to 8-bit integer precision.
For more information on the dataset, refer to
\cite{dumitru2016land}.

\section{Experiments}
\label{sec:experimnts}

In our experiments, we first set a baseline for deep learning based
SAR classification, and go on to investigate if we can improve over
that baseline with additional, synthetic data generated with a
\emph{GAN}.

\subsection{The CNN SAR classifier}

To establish a baseline for the use of CNNs with SAR data, we employ a
state-of-the-art network architecture for optical images, namely the
standard Residual Network with 50 hidden layers (\emph{ResNet-50})
\cite{he2016deep}.
To adapt the network to our class nomenclature, we remove the fully
connected layers at the top and replace them with three fully
connected layers of size $256, 256$ and $7$, respectively, which we
train from scratch. The resulting model achieves an overall accuracy
of $93.2\%$. We find this result very encouraging: in spite of the
radically different imaging process and image statistics, modern, deep
CNNs appear to be suitable for supervised SAR image classification and
yield high classification accuracy, when trained on an appropriate,
large training set.

\smallskip

\noindent
A further, interesting observation is that conventional pre-training
(i.e., initialization with the weights learned from optical images)
has little effect on the classification result.
This is not unexpected -- while the pre-training with very large
databases (millions of images) does ususally help when working with
optical images, the local image statistics of RGB and \emph{SAR} data
are probably too different to transfer even low-level image
properties.
To support that hypothesis, we have we trained the same ResNet-50
twice, once with random initialization and once with weights
pre-trained on \emph{ImageNet}. The classification results for SAR
were practical the same in both cases. I.e., the pre-trained weights
do not hurt the learning, but they also do not help compared to random
initialisation.

\subsection{Image Generation with BEGAN Models}
Given the good performance of the deep network, and the still
comparatively small training database (in computer vision, models are
routinely pre-trained with more than $10^6$ training images), we
investigate if artificial data generation with a \emph{GAN}
can further improve our classifier.
For close-range applications, it has already been shown that
classifier training can benefit from \emph{GAN} image synthesis, e.g.,
for sign recognition \cite{wang2017adversarial}.
However, our task however is more challenging, due to the extreme
variability of the SAR data in our database, and the large dimension
of the output images we need to generate ($160 \times 160$ pixels).

\subsubsection{BEGAN Model for SAR}

Despite the rather recent invention of \emph{GANs}, there is already a
plethora of variants such as \emph{DC-GANs}, \emph{cGANs},
\emph{WGANs}, \emph{DRAGANs} and \emph{BEGANs}.
We base our investigation on the newly proposed \emph{BEGAN} model
\cite{berthelot2017began}, which was shown to generate images of
remarkable quality, and to handle larger image sizes than most other
variants.
%
%

\smallskip

\noindent
Compared to the standard \emph{GAN} model, the \emph{BEGAN} design has
a number of attractive characteristics. First, it uses auto-encoders
as discriminator, thus matching the corresponding autoencoder
distributions (rather than the rawe data distributions), with a
Wasserstein distance loss.
Furthermore, \emph{BEGAN} employs an equilibrium term to balance the
effect of the \emph{Discriminator} with respect to the
\emph{Generator}, so as to avoid an ``early win'' of one stage over
the other.


\smallskip

\emph{BEGAN} was initially proposed for generating human faces. Even
though this is already a challenging problem, synthesising SAR images
proved to be a lot harder.
Through empirical experimentation, we found that the capacity of the
original model is not sufficient to capture the complexity of our
database. We therefore added more layers both to the \emph{Generator}
and the \emph{Discriminator}.
In each of the two stages, we add two additional convolution layers
(with respective eLU non-linearities), before the respective
pooling/upsampling layers.
Furthermore, we have replaced the final, linear layers of both stages
with non-linear ones, using the ReLU non-linearity.
\footnote{\scriptsize{\textbf{Code}:} \tiny{\url{https://github.com/deep-unlearn/Big_Data_From_Space_2017}}}

\smallskip

\noindent
Finally, and perhaps most significantly, we have changed the loss function of the discriminator.%
The original loss function is simply the mean of the per-pixel \emph{$L_1$} 
distance.
In our model, we replace it by a combination of a per-pixel distance
and a histogram distance, to explicitly match the global intensity
distributions of the images.
The new loss is given by :

\smallskip

	\scalebox{1.1}{%
	$\mathcal{L}_\text{generated} = L_\text{hist} + \omega \cdot L_\text{spatial} $}
	
	\smallskip
	
	\scalebox{1.1}{%
		$ L_\text{hist}=\frac{1}{N_\text{bins}} \cdot \sum (\text{hist}(X)-\text{hist}(X_\text{recon}))^2 $ 	
	}
	
	\smallskip
		
	\scalebox{1.1}{%
		$ L_\text{spatial} = \frac{1}{N_\text{pix}} \cdot \sum (X-X_\text{recon})^2 $	
	}\;,	

\smallskip

\noindent
where \emph{hist} returns the histogram of an image over a fixed
number $N_\text{bins}$ of bins (set to 64), and $N_\text{pix}$ is
the number of pixels in the generated image.
The hyperparameter $\omega$ defines a weighting between the two parts
of the loss. For our experiments we empirically set it to
$\omega=0.001$.
%

\subsubsection{BEGAN Image Generation}

Image generation with GANs still remains somewhat a brittle and
somewhat challenging task. We thus investigate three for our SAR image
generation problem. They are:

\begin{itemize}

  \vspace{-0.33em}
  \item[$\bullet$] In the hard scenario, the network is asked to
    directly generate large SAR patches of size $160 \times 160$
    pixels. This scenario would be optimal, in the sense that it
    outputs patches at the correct size for our database; but it also
    the most complex prediction task.

  \vspace{-0.33em}
  \item[$\bullet$] In the intermediate scenario, the network generates
    SAR patches at $2\times$ larger GSD, with dimension $80 \times 80$
    pixels, which are then compared to downsampled real images. The
    reduced resolution lowers the complexity of the task, while the
    patch size in scene coordinates, and thus the spatial context,
    remains the same. But the resulting images must be upsampled to
    the original dimensions, and thus lack high-frequency detail.

  \vspace{-0.33em}
  \item[$\bullet$] In the simple scenario, images are also generated
    at $80 \times 80$ pixels, but this time the original GSD is
    retained. Instead, the patch size in scene coordinates is halved,
    respectively the real SAR patches are cropped. The resulting
    images must again be upampled, to match the patch size used for
    classification. Using smaller \emph{and} more local patches
    presumably further reduces the complexity of the prediction, the
    price to pay is a mismatch in GSD between synthetic and real
    training images, and the loss of 3/4 of the context area.

\end{itemize}

\noindent
So far, we were unsuccessful in our attempts to train the hard
scenario. We leave it to future work to determine whether this can be
remedied, or whether a higher-capacity model is needed.
For the intermediate scenario, the generator appeared to converge
better, but its outputs were still unsatisfactory and did not visually
resemble the original data.
For the time being, this failure leaves us with the simple
scenario. That setting did converge to a reasonable solution that
outputs realistically-lookinf synthetic images, see examples in
\textit{Figure 1} and real SAR data in \textit{Figure
  2}.
However, one can also clearly see that the smaller patches capture
less of the context.

\subsection{Classification Augmentation Through GANs}

In spite of the limited success to synthesize full-size patches, we
continued the experiment. The ``simple''patches were upsampled to
$160\times 160$ pixels and added to the training data for the
classification network.
As a first test, we generated 5100 synthetic instances of the
\textit{Settlement} class, which is the most frequent class in the
dataset (25'000 real traning patches), and also the one
with the strongest intra-class variation.

\smallskip

\noindent
Somewhat surprisingly, retraining the \emph{ResNet-50} classifier with
the augmented dataset did not influence the classifier either way.
We get the same classification accuracy of $93.2\%$. Seemingly, the
synthetic examples were neither capable of adding any additional
information that would have improved the classifier, nor were they
unrealistic enough to negatively impact the classifier.
Obviously, in the absence of a satisfactory explanation such an
outcome appears unlikely. Future work will have to determine the
cause, and hopefully address the current short-comings of the
generator, so as to further improve the classifier network.

\section{Conclusions}

We have introduced a new, large-scale database of SAR patches with
asociated semantic class labels. To our knowledge, this is the first
SAR dataset large enough to train modern deep neural networks, and we
have demonstrated that capability by learning a \emph{ResNet-50}
convolutional network that achieves an excellent 93.2\% hit rate over
7 different scene categories.
We have further adapted the generative BEGAN network model to SAR
data, and have experimented with synthetically generated images to
obtain an even larger traning set.
Unfortunately, we are still struggling with technical problems in the
image synthesis, and the first experiments with additional, synthetic
training data have not yet led to conclusive results.
Nevertheless, our paper clearly shows that, as soon as enough data is
available, deep convolutional networks work extremely well also for
SAR images. More detailed tests and comparisons still need to be run,
but we believe that our results set a new standard for patch-wise SAR
classification.
We also posit that our failure to exploit synthetic images is due to
relatively minor technical difficulties that can be addressed, and we
are still convinced that GANs have the potential to support the the
generation of truly big training databases.

\begin{table*}[t]
	\centering
	\begin{tabular}{ccccccc}
			\includegraphics[scale=0.32]{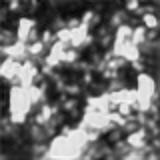}  & 
			\includegraphics[scale=0.32]{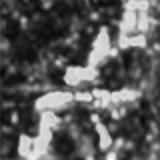}  & 
			\includegraphics[scale=0.32]{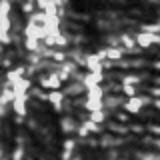}  & 
			\includegraphics[scale=0.32]{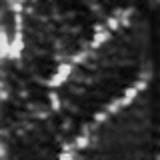}  & 
			\includegraphics[scale=0.32]{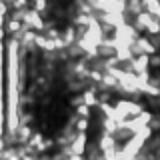}  &
   			\includegraphics[scale=0.32]{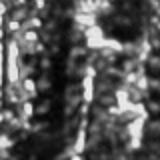}  &
   			\includegraphics[scale=0.32]{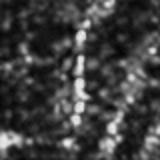}  \\
   			
   			\includegraphics[scale=0.32]{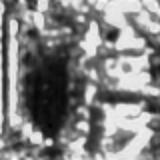}  &
   			\includegraphics[scale=0.32]{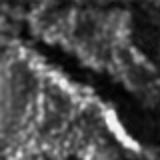}  &   			
   			\includegraphics[scale=0.32]{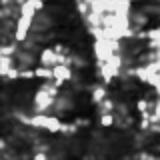}  & 
   			\includegraphics[scale=0.32]{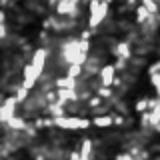}  & 
   			\includegraphics[scale=0.32]{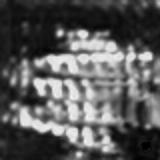}  &
   			\includegraphics[scale=0.32]{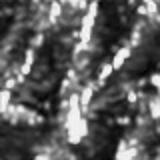}  &
   			\includegraphics[scale=0.32]{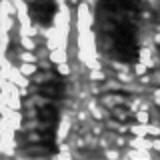}  \\
   			
	\multicolumn{7}{c}{\textbf{Figure 1.} Generated data of size $80 \times 80$ pixel by cropping scenario - upsampled to $160 \times 160$ pixel}
	\end{tabular}
	\label{generated-data}
\end{table*}

\begin{table*}[t]
	\centering
	\begin{tabular}{ccccccc}
			\includegraphics[scale=0.64]{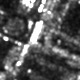}  & 
			\includegraphics[scale=0.64]{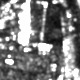}  & 
			\includegraphics[scale=0.64]{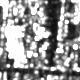}  & 
			\includegraphics[scale=0.64]{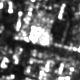}  & 
			\includegraphics[scale=0.64]{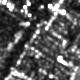}  &
   			\includegraphics[scale=0.64]{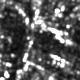}  &
   			\includegraphics[scale=0.64]{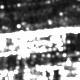}  \\
   			
   			\includegraphics[scale=0.64]{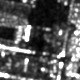}  &
   			\includegraphics[scale=0.64]{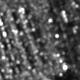}  &   			
   			\includegraphics[scale=0.64]{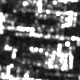}  & 
   			\includegraphics[scale=0.64]{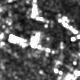}  & 
   			\includegraphics[scale=0.64]{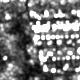}  &
   			\includegraphics[scale=0.64]{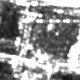}  &
   			\includegraphics[scale=0.64]{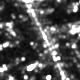}  \\
   			
	\multicolumn{7}{c}{\textbf{Figure 2.} Original \emph{TerraSAR-X} data of original size - $160 \times 160$ pixel}
	\end{tabular}
	\label{real-data}
\end{table*}



\bibliographystyle{IEEEbib}
\bibliography{strings,refs}

\end{document}